%% file: main.tex
\documentclass[runningheads]{llncs}
\usepackage[T1]{fontenc}
\usepackage{graphicx}
\title{Deep kernel video approximation for unsupervised action segmentation}
\author{S.L. Pintea\inst{1}\orcidID{0000-0002-2356-2140} \and J. Dijkstra\inst{2}\orcidID{0000-0002-8666-3731}}
\authorrunning{S.L Pintea and J. Dijkstra}
\institute{Department of Intellgient Systems (DIS), Tilburg University, Tilburg, NL \and
Division of Image Processing, LUMC, Leiden, NL
}
\graphicspath{{./img/}}
\input{macrosjvg}

\begin{document}

\maketitle

\input{paper/abstract}

\input{paper/introduction}

\input{paper/related}
\input{paper/method}
\input{paper/experiments}

\input{paper/discussion}

{
    \small
    \bibliographystyle{ieeetr}
    \bibliography{main}
}
\end{document}

%% file: macrosjvg.tex
\usepackage{multirow}
\usepackage{xcolor}

\newcommand{\slp}[1]{[{\color{green} Silvia: #1}]}

\newcommand{\Eq}[1]{Eq.~(\ref{eq:#1})}
\newcommand{\eq}[1]{\Eq{#1}}
\newcommand{\fig}[1]{Fig.~\ref{fig:#1}}
\newcommand{\tab}[1]{Tab.~\ref{tab:#1}}

\newcommand\norms[1]{\left\lVert#1\right\rVert}
\usepackage{arydshln}

\usepackage{wrapfig}
\usepackage{lipsum}  
\usepackage{graphicx}
\usepackage{amsmath}
\usepackage{amssymb}
\usepackage{booktabs}
\usepackage{comment}
\usepackage[linesnumbered,ruled]{algorithm2e}
\usepackage{algpseudocode}

\usepackage{pifont}

\def\argmax{\operatornamewithlimits{\rm arg\,max}}

\usepackage{hyperref}

\usepackage{xspace}
\newcommand{\etal}{\textsl{et~al.}\xspace}
\newcommand{\ie}{\textsl{i.e.}\xspace}
\newcommand{\etc}{\textsl{etc.}\xspace}
\newcommand{\eg}{\textsl{e.g.}\xspace}

%% file: paper/abstract.tex
\begin{abstract}
This work focuses on per-video unsupervised action segmentation, which is of interest to applications where storing large datasets is either not possible, or nor permitted.
We propose to segment videos by learning in deep kernel space, to approximate the underlying frame distribution, as closely as possible.
To define this closeness metric between the original video distribution and its approximation, we rely on maximum mean discrepancy (MMD) which is a geometry-preserving metric in distribution space, and thus gives more reliable estimates.
Moreover, unlike the commonly used optimal transport metric, MMD is both easier to optimize, and faster. 
We choose to use neural tangent kernels (NTKs) to define the kernel space where MMD operates, because of their improved descriptive power as opposed to fixed kernels.
And, also, because NTKs sidestep the trivial solution, when jointly learning the inputs (video approximation) and the kernel function. 
Finally, we show competitive results when compared to state-of-the-art per-video methods, on six standard benchmarks. 
Additionally, our method has higher $F_1$ scores than prior agglomerative work, when the number of segments is unknown.

\keywords{Unsupervised action segmentation \and MMD \and infinite NTK}
\end{abstract}

%% file: paper/introduction.tex
\section{Introduction}
\label{sec:intro}
Could you find the unitary action segments in a random video?
What if you are told beforehand how many segments there are?
This is the task of unsupervised action segmentation, where videos are segmented into mutually exclusive segments, \ie actions.
This task is essential in healthcare and assisted living, for activity tracking.
However, in these applications, storing large datasets is not permitted due to privacy concerns.
Here, we focus on such data-constrained scenarios where only one video is available at a time, with no access to large sets of training videos, and no annotations.

Prior unsupervised action segmentation works \cite{du2022fast,sarfraz2021temporally,spurio2025hierarchical,xing2024unsupervised} do not model the underlying video distribution.
Therefore, they do not take into account the underlying geometry of the video space.
In consequence, these methods do not rely on metrics of convergence in distribution space, and thus, are unstable with respect to deformations of the distribution supports \cite{feydy2019interpolating}. 
To date, two classes of commonly used distances take the geometry of the underlying distribution into account: OT (optimal transport) and MMD (maximum mean discrepancy). 
While OT has been previously used for unsupervised action segmentation \cite{ali2025joint,xu2024temporally}, it is computationally expensive and hard to optimize. 
On the other hand, MMD defines geometry-aware distances, and is more straight-forward to optimize than OT 
\cite{feydy2019interpolating}.
Moreover, MMD can be simply optimized in batches over the data.

\begin{figure}[t!]
    \centering
    \vspace{-10px}
    \includegraphics[width=.8\linewidth]{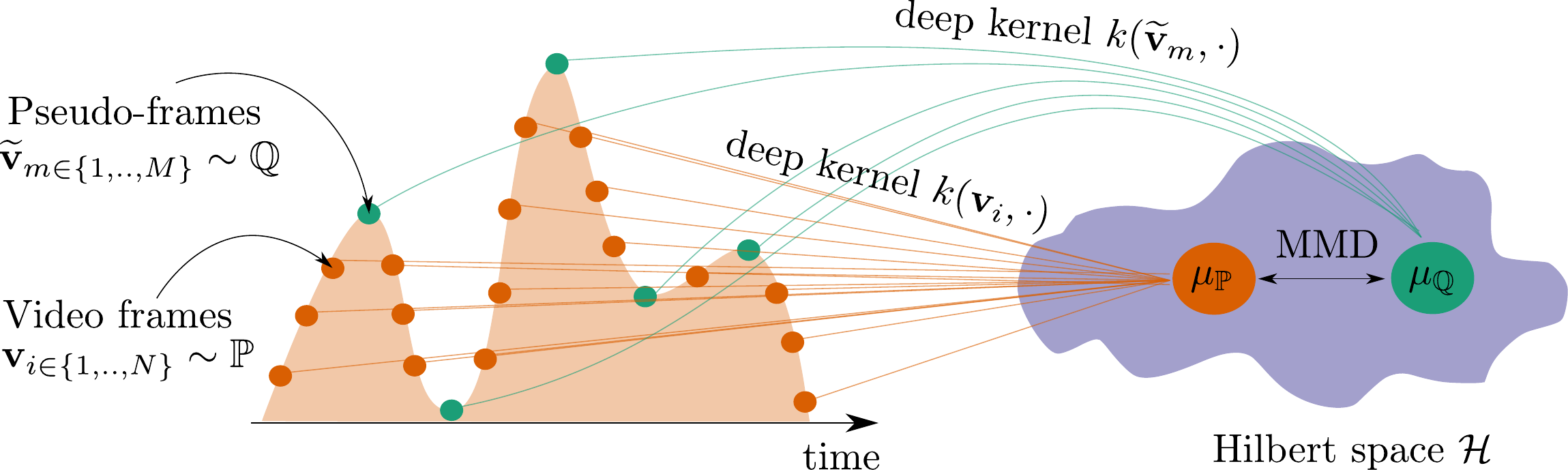}  
    \vspace{-10px}
    \caption{ \small
        We learn a small video approximation $\widetilde{\mathbf{v}}_m{\sim}\mathbb{Q}$ (in green) such that its distribution is \textsl{"as close as possible"} to
        the distribution of the real video frames $\mathbf{v}_i{\sim}\mathbb{P}$ (in orange).
        To define \textsl{"as close as possible"} we rely on a geometry preserving metric -- MMD (maximum mean discrepancy) which minimizes the distance between the kernel means $\mu_\mathbb{Q}$ and $\mu_\mathbb{P}$.
    }
    \label{fig:intro}
\end{figure}


This work proposes learning a video approximation in a deep kernel formulation by optimizing a geometry-preserving metric, namely MMD. 
By definition, MMD works in kernel space \cite{gretton2006kernel}.
Although using fixed kernels (such as defined by the exponential family) is better suited for cases such as ours where we do not rely on a training set, these kernels are limited in descriptive power. 
Wenliang \etal \cite{wenliang2019learning} show that deep network kernels are more informative than the exponential-family kernels.
Here, we want to get the best of both worlds, by choosing to use NTKs (neural tangent kernels). 
NTKs combine the non-parametric flexibility of kernel methods, with the descriptive power of deep neural networks. 
Specifically, here we rely on infinite NTKs over fully-connected networks, which consistently outperform finite NTKs \cite{lee2020finite}.
Concretely, our method learns to approximate a set of video frames $\{\mathbf{v}_i\}_{i\in \{1,..,N\}}$ by a small video approximation $\{\widetilde{\mathbf{v}}_m\}_{m\in\{1,..,M\}}$ with $M{\ll}N$.
Our goal is to learn a distribution over the video approximation $\widetilde{\mathbf{v}}{\sim}\mathbb{Q}$ that is as \textsl{close as possible} to the true video distribution $\mathbf{v}{\sim}\mathbb{P}$. 
To define \textsl{``as close as possible"} we minimize a geometry-preserving metric between the real video distribution and its approximation -- MMD, as in \fig{intro}.

To summarize, this work is the first to propose learning a video approximation in a deep kernel formulation by optimizing via a geometry-preserving metric -- MMD (maximum mean discrepancy).
Specifically,
(i) we focus on per-video unsupervised action segmentation where we learn a small set of synthetic frames that closely approximate the original video; 
(ii) to this end, we aim to make the distribution of the video approximation as close as possible to the real video frames distribution;
(iii) we show experimentally that the proposed method is more precise in terms of $F_1$ scores, than prior agglomerative work, when the number of segments is not known; and our method has competitive results when benchmarked on standard datasets.
Our complete code is available at: \href{https://github.com/SilviaLauraPintea/video_approx}{https://github.com/SilviaLauraPintea/video\_approx}. 


%% file: paper/related.tex
\section{Related work}
For an overview of the state-of-the-art action segmentation, we direct the readers to \cite{ding2023temporal}.
Here, we focus on unsupervised temporal segmentation of actions. 

\medskip\noindent\textbf{Clustering-based action segmentation.}
Unsupervised methods rely on probabilistic mixture models \cite{kitani2011fast,nater2011temporal,sener2018unsupervised}. 
Using a stacked Dirichlet mixture process \cite{kitani2011fast} pioneers unsupervised action segmentation. 
Similarly, a mixture of Gaussians can be used for coarse-to-fine clustering of frames \cite{nater2011temporal}, or over U-Net embeddings \cite{vidalmata2021joint}.
Also from a probabilistic perspective, a Generalized Mallows Model \cite{sener2018unsupervised} is used to alternate between learning action appearance and temporal modeling.
A parallel research trend has focused on clustering self-supervised features \cite{dimiccoli2020learning,kukleva2019unsupervised,wang2022sscap}.
Rather than clustering embeddings, learning hierarchical codebooks in an encoder-decoder architecture is effective \cite{spurio2025hierarchical}.  
Graph embeddings \cite{dimiccoli2020learning} or simple MLP embeddings \cite{kukleva2019unsupervised} can be clustered.  
Vice versa, starting from a given video clustering, it is effective to further group frames based on visual similarity \cite{bueno2023leveraging}.
While network embeddings are informative, a training-free temporally-weighted hierarchical clustering \cite{sarfraz2021temporally} is both effective and fast.
We are dissimilar to the above methods, as our video approximation is not defined by cluster centers, but learned, such that it follows the underlying video-frame distribution.

\medskip\noindent\textbf{Action segmentation by labeling.}
Rather than clustering, detecting action boundaries based on frame similarities also gives promising results \cite{du2022fast,xing2024unsupervised}.
Action boundaries can be detected as discrepancies between predicted future frames and true future frames \cite{aakur2019perceptual,garcia2018predicting}.
Alternatively, an auto-encoder with an attention module trained using a contrastive loss is effective at finding discriminative latent codes \cite{swetha2021unsupervised}.
Differently, generating labels and recovering the action ordering through expectation-maximization over HMM (Hidden Markov Model) \cite{li2021action} or OT (optimal transport) \cite{ali2025joint,kumar2022unsupervised,xu2024temporally} has been previously proposed.
Rather than generating labels per frames using self-supervision or optimal transport, here, we aim to learn a video approximation in deep kernel space. 
For this, our method relies on a geometry preserving metric --- MMD (maximum mean discrepancy), which unlike OT, has improved computational and sample complexity and does not suffer from the curse of dimensionality. 
Moreover, because we train per video rather than over a complete dataset of videos, we opt for a computationally efficient deep kernel model -- an infinite NTK (neural tangent kernel). 

\medskip\noindent\textbf{Learning pseudo-inputs.}
Approximating a large dataset with a small set of pseudo-samples, called \textsl{inducing points}, has been studied in the Gaussian Process literature \cite{csato2002sparse,smola2000sparse,snelson2005sparse}.
More recently, this idea has been adopted in dataset distillation literature \cite{nguyen2020dataset} in combination with NTKs (neural tangent kernels) \cite{jacot2018neural}.
Kernel ridge regression using a kernel defined as NTK, approximates infinitely wide neural networks \cite{arora2019exact,jacot2018neural}.
Here, rather than learning the video approximation in a supervised manner in a Gaussian Process formulation, we rely on MMD (maximum mean discrepancy). 
MMD in kernel space was first defined in \cite{gretton2006kernel,smola2006maximum}, to test if two groups of samples come from the same distribution. 
More recently, NTKs have been extended to approximate characteristic kernels, and thus be useful for hypothesis testing \cite{cheng2021neural,geifman2020similarity,jia2021efficient}.
We build on prior work, and use infinite NTKs \cite{lee2020finite} to learn a video approximation that minimizes the MMD with respect to the real video frame distribution. 

%% file: paper/method.tex
\section{Learning video approximations in deep kernel space}
Given a set of video frames ${\mathbf{v}}_{i\in\{1,..,N\}}$, we want to learn a small set of synthetic frames ${\widetilde{\mathbf{v}}}_{m \in \{1,..,M\},~ M\ll N}$ that closely approximate the video. 
During the optimization, to objectively measure the closeness between the original video and its approximation, we need a metric.
Here, we opt for a geometry-preserving metric, specifically MMD (maximum mean discrepancy).

\medskip\noindent\textbf{Geometry preserving metric: MMD.}
We want the distribution of the video approximation $\widetilde{\mathbf{v}}$ to be as close as possible to the distribution of the real video $\mathbf{v}$.
Specifically, we assume that the real video frames are drawn from $\mathbb{P}$: $\mathbf{v} {\sim} \mathbb{P}$, and the video approximation from $\mathbb{Q}$: $\widetilde{\mathbf{v}} {\sim} \mathbb{Q}$.  
And we optimize $\widetilde{\mathbf{v}}$ such that $\mathbb{Q} \approx \mathbb{P}$.

By definition, MMD \cite{gretton2006kernel} operates in the RKHS (Reproducing Kernel Hilbert Space) \cite{gretton2006kernel} denoted by $\mathcal{H}$.
In RKHS, starting from a mapping function $k_\mathbf{v}{:}\mathcal{V}{\rightarrow}\mathcal{H}$, $k_\mathbf{v}(\cdot){=}k(\mathbf{v}, \cdot)$, we can define a kernel function $k{:}\mathcal{V}{\times}\mathcal{V}{\rightarrow}\mathbb{R}$ as an inner product $k(\mathbf{v}_i,\mathbf{v}_j) {=} \langle k(\mathbf{v}_i,\cdot), k(\mathbf{v}_j, \cdot)\rangle$.
In RKHS we can measure the distance between two distributions by measuring the distance between their mean embeddings \cite{gretton2006kernel}: 
\begin{alignat}{1}
    \text{MMD} & (\mathbb{P}, \mathbb{Q}, \mathcal{H}) {=} \text{sup}_{k{\in}\mathcal{H}} \left( \mathbf{E}_{\mathbf{v}{\sim}\mathbb{P}} [k(\mathbf{v},\cdot)] {-}\mathbf{E}_{\widetilde{\mathbf{v}}{\sim}\mathbb{Q}} [k(\widetilde{\mathbf{v}},\cdot)] \right)
\end{alignat}
The empirical estimate of MMD is given by the distance between the kernel means, estimated over samples from the data distributions $\mathbb{P}$ and $\mathbb{Q}$:   
\begin{alignat}{1}
    \text{MMD} & (\mathbb{P}, \mathbb{Q}, \mathcal{H}) =\norms{ \boldsymbol{\mu}_\mathbb{P} - \boldsymbol{\mu}_\mathbb{Q}}_\mathcal{H} = \norms{ \frac{1}{N}\sum_{\mathbf{v} \sim \mathbb{P}}  k(\mathbf{v},\cdot) - \frac{1}{M}\sum_{\widetilde{\mathbf{v}}\sim \mathbb{Q}} k(\widetilde{\mathbf{v}},\cdot) }_\mathcal{H}
    \label{eq:mmd}
\end{alignat}
Gretton \etal \cite{gretton2006kernel} show that this can be rewritten in a simple form that only depends on kernel values:
\begin{alignat}{1}
\text{MMD}^2 & (\mathbb{P}, \mathbb{Q}, \mathcal{H}) {=} \frac{1}{N^2}\sum_{i,j}  k(\mathbf{v}_i,\mathbf{v}_j) {+} \frac{1}{M^2}\sum_{n,m} k(\widetilde{\mathbf{v}}_n,\widetilde{\mathbf{v}}_m) {-} \frac{2}{N M}\sum_{i,m} k(\mathbf{v}_i,\widetilde{\mathbf{v}}_m) 
    \label{eq:mmd2}
\end{alignat}
MMD has the nice property that it does not need to make any assumption about the underlying data distributions, $\mathbb{P}$ and $\mathbb{Q}$. 
However, we note here that $MMD$, as defined in \cite{gretton2006kernel}, assumes that the samples $\mathbf{v}{\sim}\mathbb{P}$ and $\widetilde{\mathbf{v}}{\sim}\mathbb{Q}$ are i.i.d samples.
While this assumption may hold for unitary action prototypes $\widetilde{\mathbf{v}}$, this does not hold for video frames, $\mathbf{v}$.
To compensate for this, we learn the video approximation $\widetilde{\mathbf{v}}$ by minimizing the \eq{mmd2} using shuffled batches of the data of size $\frac{N}{M}$.
Additionally, we regularize the norm of $\widetilde{\textbf{v}}_m$, to constrain the learning. 

Ideally, when optimizing $MMD^2(\mathbb{P}, \mathbb{Q}, \mathcal{H})$, the kernels should satisfy two other properties: \ie that the mapping $k(\mathbf{v}_i,\cdot)$ is smooth, and that the kernel is \textsl{characteristic}.
For mapping smoothness, we use kernels defined by continuous functions, discussed below.
For the second property, this depends on the kernel choice. 
A \textsl{characteristic} kernel has a mean embedding that fully describes the data distribution.
Therefore, for these kernels, it holds that when $MMD(\mathbb{P}, \mathbb{Q}, \mathcal{H}){=}0$ this implies $\mathbb{P}{=}\mathbb{Q}$.
However, the standard characteristic kernels (\eg the exponential family) have limited representation power \cite{wenliang2019learning}.
Therefore, we choose to combine an exponential kernel with a deep learning kernel. 

\begin{wrapfigure}{r}{0.5\textwidth}
    \centering
    \vspace{-20px}
    \includegraphics[width=1\linewidth]{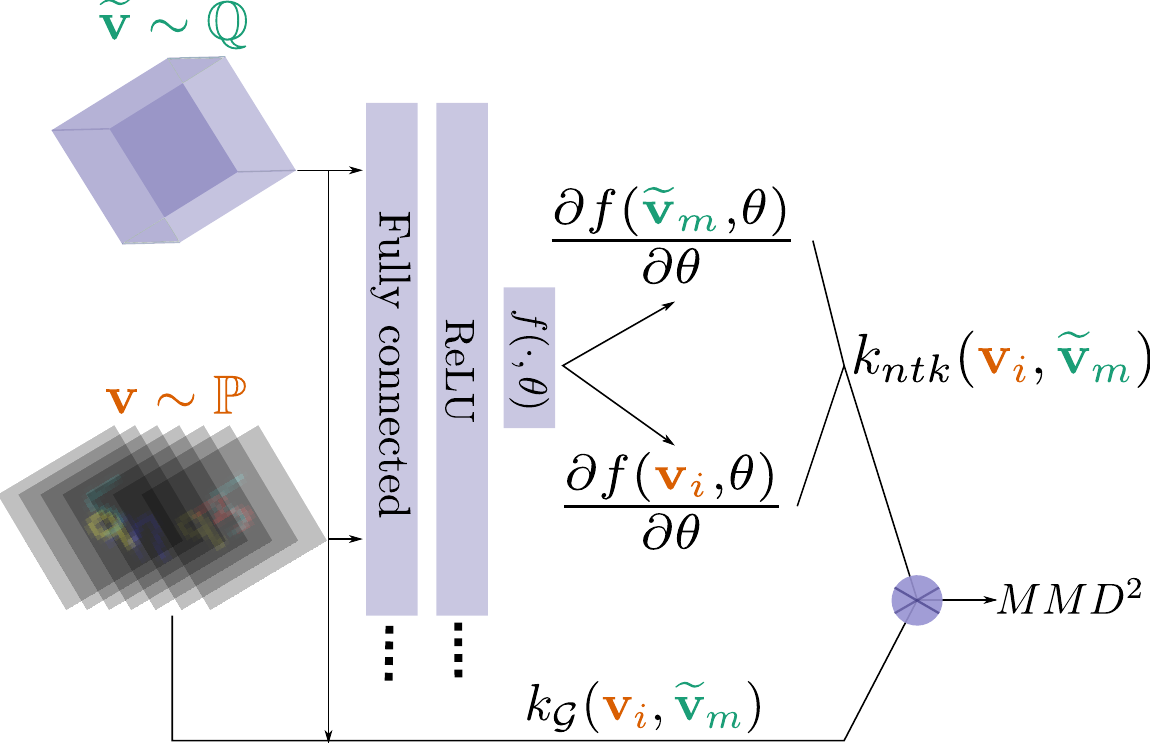}  
    \vspace{-20px}
    \caption{\small
        We learn the video approximation $\widetilde{\mathbf{v}}$ by optimizing the $MMD^2$ loss 
        over a product of NTK and Gaussian kernels.
    }
    \vspace{-15px}
    \label{fig:ntk}
\end{wrapfigure}
\medskip\noindent\textbf{Non-parametric yet descriptive deep kernels: infinite-NTKs.}
MMD relies on kernel functions $k(\cdot, \cdot)$. 
Learning the kernel function $k(\cdot, \cdot)$ together with the video approximation $\widetilde{\mathbf{v}}$ leads to a trivial solution.
Therefore, we need to rely on fixed kernels, such as the exponential family kernels. 
However, Wenliang \etal \cite{wenliang2019learning}, shows that the exponential family kernels are less informative than deep network kernels.
Thus, to have a fixed kernel and yet a deep kernel, we choose to use NTKs (neural tangent kernels).
NTKs \cite{smola2000sparse,snelson2005sparse} are kernels defined over the gradients of wide deep networks. 
Particularly, we choose to use infinite NTKs, which consistently outperform finite NTKs \cite{lee2020finite}.

The NTK, at two samples $\mathbf{v}_i,\mathbf{v}_j$ for a neural network $f(\cdot,\boldsymbol{\theta})$, is the sum over layers $l$ of inner products between the gradients of the network output with respect to the layer parameters $\boldsymbol{\theta}_l$, evaluated at samples $\mathbf{v}_i,\mathbf{v}_j$:
\begin{alignat}{1}
    k_{ntk}(\mathbf{v}_i,\mathbf{v}_j) = \sum_l \Bigg\langle \ \frac{\partial f(\mathbf{v}_i,\boldsymbol{\theta})}{\partial \boldsymbol{\theta}_l} \frac{f(\mathbf{v}_j,\boldsymbol{\theta})}{\partial \boldsymbol{\theta}_l} \Bigg\rangle.
    \label{eq:ntk}
\end{alignat}

The NTK in \eq{ntk} is not a characteristic kernel. 
To ameliorate this, we combine the $k_{ntk}(\cdot, \cdot)$ with a characteristic kernel, namely a Gaussian kernel: $k_\mathcal{G}(\mathbf{v}_i,\mathbf{v}_j)$.
This combination has the added benefit that it allows us to minimize the distance between the distributions both in input image space (\ie Gaussian kernel) and in the neural network gradient space (\ie NTK).
To combine the kernels, we simply take the product.
Because the two kernels have different ranges, we define a rescaling $\alpha {=} \frac{\text{med}\left(k_\mathcal{G}(\mathbf{v}_i,\mathbf{v}_j)\right)}{\text{med}\left(k_{ntk}(\mathbf{v}_i,\mathbf{v}_j)\right)}$, where $\text{med}(\cdot)$ denotes the median, to bring them in the same range:
\begin{alignat}{1}
    k(\mathbf{v}_i,\mathbf{v}_j) = \alpha k_{ntk}(\mathbf{v}_i,\mathbf{v}_j) k_\mathcal{G}(\mathbf{v}_i,\mathbf{v}_j),
    \label{eq:kern}
\end{alignat}
where $k_\mathcal{G}(\mathbf{v}_i, \mathbf{v}_j) {=} \text{exp}\left(- \frac{\norms{\mathbf{v}_i - \mathbf{v}_j}_2^2}{\lambda^2} \right)$ is the Gaussian kernel with length-scale $\lambda$.
Note that when the length-scale $\lambda{\rightarrow}\infty$, the Gaussian kernel $k_\mathcal{G}(\cdot,\cdot){\rightarrow}0$.
Therefore, jointly learning the length-scale $\lambda$ and the $\widetilde{\mathbf{v}}$, leads to a trivial solution.
To avoid this, inspired by Gretton \etal \cite{gretton2012kernel}, we set $\lambda = \text{med}_{i,j} (\norms{\mathbf{v}_j-\mathbf{v}_i}_2^2)$, where $\text{med}(\cdot)$ is the median function.
Rather than using the median of the distances as in \cite{gretton2012kernel}, we use the median of the squared distances, as it works better in combination with an NTK, and acts as a regularization that enforces smoothness. 
\fig{ntk} shows a schematic representation of our video approximation learning.

Our overall kernel in \eq{kern} is not a characteristic kernel, as $k_{ntk}(\cdot, \cdot)$ is not characteristic \cite{jia2021efficient}.  
Projecting the inputs on a hypersphere $\mathbb{S}^{d-1}$ where $d$ is the dimension of the inputs can make NTK similar to a Laplace kernel \cite{geifman2020similarity}, which is characteristic.
Alternatively, using random cosine projections in the first layer \cite{jia2021efficient} also makes NTK approximate a characteristic kernel.
In the experimental section we analyze these kernel alternatives.
  
\medskip\noindent\textbf{Action segmentation.}
We use the kernel in \eq{kern} and learn a video approximation that minimizes the distribution distance in kernel space, as defined by the $MMD^2(\mathbb{P}, \mathbb{Q}, \mathcal{H})$ metric, between the real video $\mathbf{v}$ and its approximation $\widetilde{\mathbf{v}}$.
Given that the kernel can be interpreted as a similarity between samples, we use this to assign the real video frames to the synthetic frames in the approximation, and thus segment the actions.  
Concretely, we assign a video frame $\mathbf{v}_i$ to an action indexed by $m^*$, by taking the maximum kernel similarity across all $\widetilde{\mathbf{v}}_m$:
\begin{alignat}{1}
    m^* = \argmax_m  k(\mathbf{v}_i, \widetilde{\mathbf{v}}_m).
    \label{eq:align}
\end{alignat}
Following prior work \cite{sarfraz2021temporally,du2022fast,xu2024temporally}, we use Hungarian matching to map the indexes $m{\in}\{1,.., M\}$ to the ground truth action classes in the dataset.
In our method, aligning the video approximation with the original video causes severe over-segmentation when the features are not smoothed over time.
Therefore, we smooth the input video frames using a Gaussian, and discuss in the experiments the impact of the smoothing hyperparameter.


\medskip\noindent\textbf{Theoretical time\slash complexity analysis.}
Kernel mean embeddings, as used in the MMD (\eq{mmd2}) have the benefit that they can model higher order moments of the distributions, while being simple to optimize and fast \cite{feydy2019interpolating,muandet2017kernel}.
When compared to optimal transport -- an alternative geometry-preserving metric, MMD is provably more efficient. 
MMD\rq s computational complexity is $\mathcal{O}(n^2)$ \cite{gretton2006kernel} while for optimal transport this is $\mathcal{O}(n^3 \log(n))$ \cite{genevay2019sample}, where $n$ is the number of samples.
Additionally, MMD is independent of data dimension and has a sample complexity of $\mathcal{O}\left(\frac{1}{\sqrt{n}}\right)$, while optimal transport suffers from the curse of dimensionality with a sample complexity of $\mathcal{O}\left(n^{-\frac{1}{d}}\right)$ \cite{genevay2019sample}, where $d$ is the data dimension. 
In practice, we can improve the speed of MMD by optimizing it over batches of data.
Thus, allowing us to process extremely long videos.
Moreover, using batches gives better results than optimizing over the complete video.

%% file: paper/experiments.tex
\section{Experimental analysis}

\noindent\textbf{Datasets overview.}
We compare with state-of-the-art on six standard segmentation benchmarks: \textsl{Breakfast} \cite{kuehne2014language}, \textsl{50 Salads} \cite{stein2013combining}, \textsl{YTI (YouTube Instructional)} \cite{alayrac2016unsupervised}, \textsl{Desktop Assembly} \cite{kumar2022unsupervised}, \textsl{Hollywood Extended} \cite{bojanowski2014weakly} and \textsl{MPII Cooking 2} \cite{rohrbach2016recognizing}.
We follow previous work \cite{du2022fast,sarfraz2021temporally,xu2024temporally}, and for \textsl{Breakfast} \cite{kuehne2014language}, \textsl{50 Salads} \cite{stein2013combining}, \textsl{Hollywood Extended} \cite{bojanowski2014weakly} and \textsl{MPII Cooking 2} \cite{rohrbach2016recognizing} we use 64-dimensional IDT (improved dense trajectory) features \cite{wang2013action}.
Also as in previous work, we use 3000-dimensional features provided by \cite{kukleva2019unsupervised} for \textsl{YTI} \cite{alayrac2016unsupervised}, while for \textsl{Desktop Assembly} \cite{kumar2022unsupervised} we use 512-dimensional features provided by \cite{kumar2022unsupervised}.
\tab{data} shows an overview of the datasets.
We do not exclude any action class during training.
On \textsl{YTI}, at evaluation time we ignore the background class, as in \cite{du2022fast,xu2024temporally}.
\begin{table}[t]
    \centering
    \resizebox{.95\linewidth}{!}{
    \setlength{\tabcolsep}{6pt}
            \begin{tabular}{l llll}
                \toprule
                & Avg. frames  & Videos & Avg. segments     & Features \\ \midrule
                50 Salads \cite{stein2013combining}           & 11788 & 50      &  19           & 64$D$ IDT \cite{wang2013action}   \\ 
                MPII Cooking 2 \cite{rohrbach2016recognizing} & 10555 & 273     &  17           & 64$D$ IDT \cite{wang2013action}   \\
                Breakfast \cite{kuehne2014language}           & 2099 & 1712    &  6            & 64$D$ IDT \cite{wang2013action}    \\
                Hollywood extended \cite{bojanowski2014weakly} & 835 & 937     &  3            & 64$D$ IDT \cite{wang2013action}               \\
                Desktop assembly \cite{kumar2022unsupervised} & 779 & 76      &  23           & 512$D$-features \cite{kumar2022unsupervised}  \\
                YTI \cite{alayrac2016unsupervised}            & 520  & 150     &  9            & 3k $D$-features \cite{kukleva2019unsupervised} \\
                \bottomrule
            \end{tabular}}
    \caption{\small
    \textbf{Datasets overview.}
    We use the same features as prior work \cite{du2022fast,sarfraz2021temporally,xu2024temporally}. 
    }
    \label{tab:data}
    \vspace{-20px}
\end{table}

\medskip\noindent\textbf{Metrics and baselines.}
Our method computes a video approximation, and aligns each video with its learned approximation.
Therefore, we can only compare with unsupervised methods that perform the segmentation per video, rather than across videos, such as: TW-FINCH \cite{sarfraz2021temporally}, ABD \cite{du2022fast} and ASOT \cite{xu2024temporally} and CLOT \cite{bueno2025clot}.
Unfortunately, we cannot compare with the recent HVQ \cite{spurio2025hierarchical} and VASOT \cite{ali2025joint}, as they use the full dataset to train the segmentation, rather than individual videos.
To compare with the baselines, we use the standard metrics: \textsl{MoF} (mean accuracy over frames); \textsl{IoU} (intersection over union) and F$_1$ from \cite{sarfraz2021temporally}.

\medskip\noindent\textbf{Hyperparameters.}
In all our experiments we use the NTK initialization with hyperparameters: $\mathbf{W}{\sim}\mathcal{N}(0, {2})$, $\mathbf{b}{\sim}\mathcal{N}(0, 0.1)$ (where $\mathbf{W},\mathbf{b}$ represent the MLP weights and biases, respectively) \cite{yang2020feature,sohl2020infinite}.
And we use a learning rate set to $5e{-}2$ and a weight decay of $1e{-}3$, for all datasets.  
We use an infinite-width NTK, defined by a 1-hidden layer MLP with ReLU activations implemented with the \textsl{neural tangents} library \cite{neuraltangents2020} in \textsl{JAX}.
We initialize the video approximation with the mean frames of uniform segments. 
For each video, we set the number of segments $M$ to the average number of actions per activity\slash dataset, as in previous work \cite{sarfraz2021temporally,xu2024temporally}.
We minimize the $MMD^2$ over 100 epochs, and use a batch size of $\frac{N}{M}$, where $N$ is the number of video frames. 
Because we perform the optimization over data batches, our method can be effectively applied to long videos. 
We ran all experiments on a small NVIDIA RTX A1000 Laptop GPU.

\subsection{Hypothesis: the video approximation learns actions}
\begin{table}[t!]
    \centering
    \begin{tabular}{lr}
            \resizebox{.5\linewidth}{!}{
            \setlength{\tabcolsep}{4pt}
            \begin{tabular}{l ccc}
                \toprule
                                            & MoF ($\uparrow$) & IoU ($\uparrow$) & $F_1$ ($\uparrow$) \\ \midrule
                K-means                       & 59.2\% & 42.2\% & 54.9\% \\ 
                Uniform                       & 68.2\% & 54.8\% & 67.3\%  \\ \midrule
                Kernel (K-means)        & 60.4\%  & 44.0\% & 56.6\%  \\ 
                Kernel (Uniform)        & 72.5\% 	& 59.3\% & 70.3\% \\
                \textsl{Video approx.} (Ours) & 73.3\% 	& 59.9\% & 70.5\% \\
                \bottomrule
            \end{tabular}}
            &
        \resizebox{.5\linewidth}{!}{
        \begin{tabular}{ccccc}
                \multicolumn{5}{c}{\large Ground truth average frame}\\
                \includegraphics[width=0.2\linewidth,decodearray={0.3 1 0.3 1 0.3 1}]{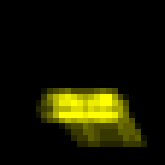} &
                \includegraphics[width=0.2\linewidth,decodearray={0.3 1 0.3 1 0.3 1}]{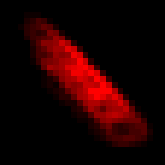} & 
                \includegraphics[width=0.2\linewidth,decodearray={0.3 1 0.3 1 0.3 1}]{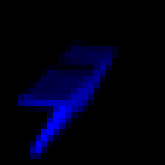} &
                \includegraphics[width=0.2\linewidth,decodearray={0.3 1 0.3 1 0.3 1}]{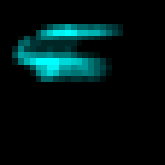} & 
                \includegraphics[width=0.2\linewidth,decodearray={0.3 1 0.3 1 0.3 1}]{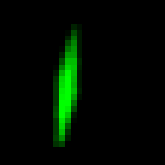} \\
                \multicolumn{5}{c}{\large \textsl{Video approximation}}\\
                \includegraphics[width=0.2\linewidth,decodearray={0.3 1 0.3 1 0.3 1}]{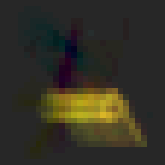} &
                \includegraphics[width=0.2\linewidth,decodearray={0.3 1 0.3 1 0.3 1}]{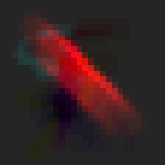} &
                \includegraphics[width=0.2\linewidth,decodearray={0.3 1 0.3 1 0.3 1}]{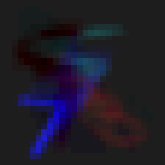} &
                \includegraphics[width=0.2\linewidth,decodearray={0.3 1 0.3 1 0.3 1}]{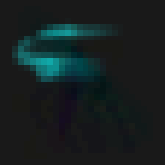} &
                \includegraphics[width=0.2\linewidth,decodearray={0.3 1 0.3 1 0.3 1}]{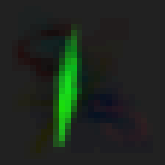} \\
            \end{tabular}}
            \\
    \end{tabular}
    \caption{\small
    \textbf{Hypothesis testing.}
        \textsl{Left:} Aligning the video approximation in kernel space improves over using $L_2$ distances for aligning cluster centers with video frames.
        \textsl{Right:} Examples of learned video approximations, versus ground truth average frame per class. 
        The learned video approximation aligns better with the video than the uniform segment centers, or K-means centers.
    }
    \vspace{-20px}    
    \label{tab:h1}
\end{table}

\begin{table}[t]
    \centering
    \begin{tabular}{c}
            \resizebox{.9\linewidth}{!}{
            \setlength{\tabcolsep}{2pt}
            \begin{tabular}{lcc lcc}
                \toprule
                & \multicolumn{2}{c}{True \# actions} & & \multicolumn{2}{c}{Random \# actions}\\
                \cmidrule(lr){2-3}
                \cmidrule(lr){5-6}
                & MoF ($\uparrow$)    & Boundary acc. ($\uparrow$) & & MoF ($\uparrow$)    & Boundary acc. ($\uparrow$) \\ \midrule
                TW-FINCH \cite{sarfraz2021temporally}    & 72.5\%  & 64.4\%
                &  TW-FINCH \cite{sarfraz2021temporally} & 60.3\%  & 58.8\%   \\
                Ours        & 73.3\%    & 51.2\%                      
                &  Ours     & 60.5\% 	& 62.4\% \\
                \bottomrule
            \end{tabular}} \\
            (a) Segmentation with true classes and random number of classes \\[5px]
            \includegraphics[width=1\linewidth, decodearray={0 .8 0 .8 0 .8}]{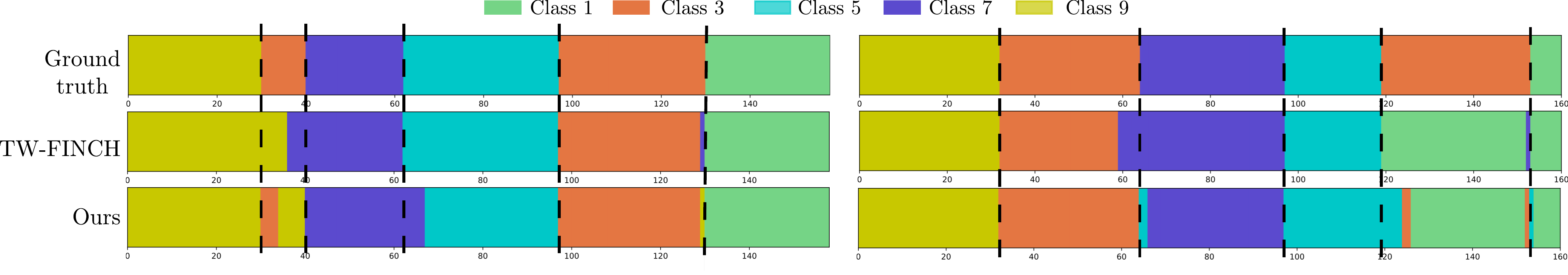} \\
            (b) True number of segments: 5$^{th}$ video (left) and 35$^{th}$ video (right).\\
            \includegraphics[width=1\linewidth, decodearray={0 .8 0 .8 0 .8}]{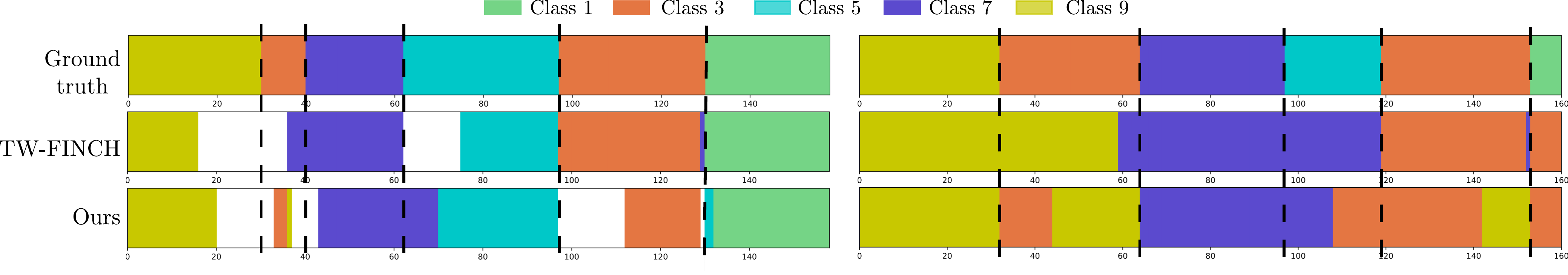} \\
            (c) Random number of segments: more (5$^{th}$ video) and less (35$^{th}$ video).\\
    \end{tabular}
    \caption{\small
    \textbf{Hypothesis testing.}
        (a) While \textsl{TW-FINCH}\cite{sarfraz2021temporally} can give precise boundaries when knowing the ground truth number of segments, it fails to do so for a random number of segments.
        (b) Two examples when predicting the true number of segments.
        (c) Tow examples when the desired number of segments is randomly varied.
        Our model can better predict repeated actions, and can better detect the boundaries.
    }
    \vspace{-20px}
    \label{tab:h12}
\end{table}

\noindent\textbf{Experimental setup.} To evaluate our hypothesis that our method can learn good video approximations, we use a controlled setting.
Concretely, we design a \textsl{Moving5} dataset containing 50 videos per set (training\slash validation\slash test) and 5 classes, where videos contain ${\approx}150$ frames. 
We select the digits $\{1,3,5,7,9\}$ to represent actions. 
Each digit moves for a variable number of frames and with a speed proportional to the number of frames, across a fixed trajectory: \ie top-to-bottom, diagonal, inverse-diagonal, right-to-left, left-to-right.    
To make it challenging, action \textsl{"3"} can repeat up to 3 times per video.
To visualize the learned video approximation, we define $\widetilde{\mathbf{v}}$ in the RGB image domain of size $M{\times}28{\times}28{\times}3$.
We normalize each frame by its $L_2$ norm.
We do not smooth the videos over time.
To avoid overfitting, we train for 10 epochs.
We select these hyperparameters on a separate validation set.

\medskip\noindent\textbf{Is learning the video approximation effective?}
\tab{h1} shows that aligning videos with their approximation, in the kernel space (using \eq{align}) is more effective than using $L_2$ distances.
Each video segment length can vary between 5 and 30 frames. 
Thus, the uniform video splitting has an error proportional to this variance.
While fixing the video approximation to uniform initialization works well, learning the video approximation further improves the results.

\medskip\noindent\textbf{Finding unitary action boundaries.}
Here, we test if our method can find unitary action boundaries.
For this, we predict a number of segments randomly varied with an offset $\delta${=}$M{\pm}\delta$, where $M$ is the true number of actions, and $\delta \in \{1, 2, ..,5\}$.
Because the data is noisy: \ie the required number of segments does not match the underlying data, we use 20 epochs here and a weight decay of $1e{-}4$.
We evaluate with respect to the true number of segments.
We estimate the boundary accuracy as the percentage of action boundaries correctly detected.
We compare with \textsl{TW-FINCH} \cite{sarfraz2021temporally}, as a representative of agglomerative clustering.

\tab{h12} shows that our method can better detect action transitions than agglomerative clustering when the required number of segments does not match the underlying data structure.
Moreover, unlike agglomerative clustering methods \cite{sarfraz2021temporally,du2022fast}, our method can deal with repeated actions, as it does not explicitly use the time information for clustering.
Our advantage is due to our method not strictly enforcing a given number of segments: although we learn $M$ synthetic frames in our video approximation, some of them may not align well with any video frame in kernel space (using \eq{align}). 

\vspace{-10px}
\subsection{Analysis of model choices}
Given that we use infinite NTK, there are no kernel parameters that need to be selected.
For the initialization of the underlying MLP weights, there is already extensive research, and we follow the standard NTK initialization \cite{sohl2020infinite}.
The actual seed used plays no role, as the networks are infinitely wide.
And, for the Gaussian kernel, we use the median of the squared distances between the samples. 
Here, we consider the effect of the kernel choice and the data smoothing.

\begin{figure}[t]
\centering
    \begin{tabular}{cc}
    \resizebox{.5\linewidth}{!}{
    \setlength{\tabcolsep}{3pt}
    \begin{tabular}{l ccc}
                \toprule
                 & MoF ($\uparrow$) & IoU ($\uparrow$) & $F_1$ ($\uparrow$)\\ \midrule
   Gauss                                   & 62.1\% & 46.0\% &	56.7\% \\
   NNGP \cite{lee2018deep}                 & 67.7\% & 46.2\% & 56.6\% \\ 
   NTK \cite{arora2019exact}               & 69.1\% & 42.3\% &	52.6\% \\ 
   NTK sphere \cite{geifman2020similarity} & 67.1\% & 46.6\% &	57.3\% \\
   SNTK \cite{jia2021efficient} \footnotemark[1] & 62.7\% & 45.7\% &	56.5\% \\ 
   Gauss${\times}$NTK sphere               & 69.8\% & 44.6\% & 55.0\%  \\
   Gauss${\times}$NTK (Ours)               & 70.1\% & 44.9\% & 55.3\%	\\ 
    \bottomrule
    \end{tabular}} &
    \begin{tabular}{c}
    \includegraphics[width=.5\linewidth, decodearray={0.5 1 0.5 1 0.5 1}]{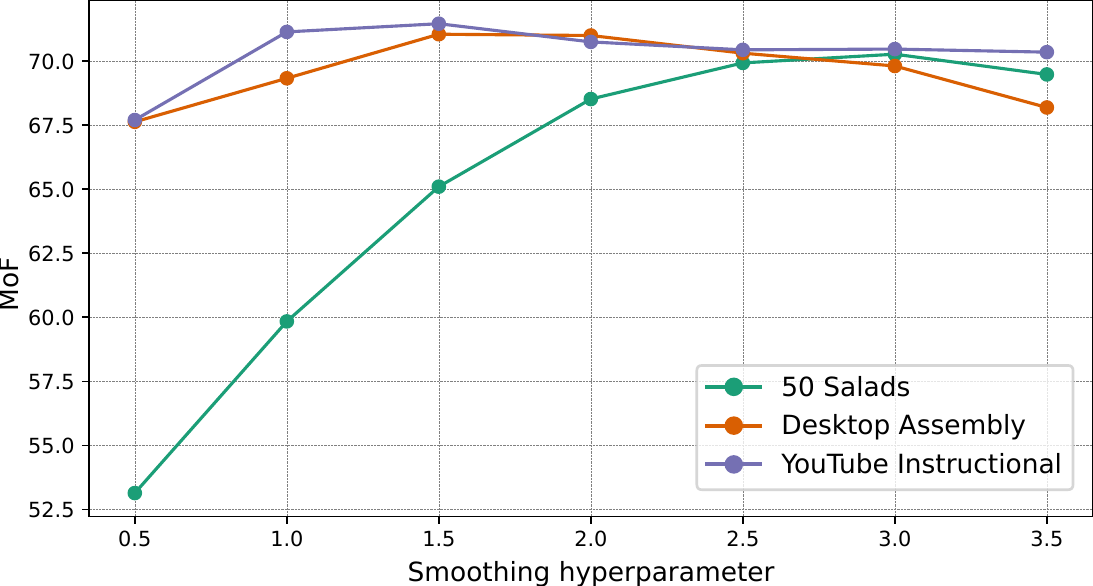} 
    \end{tabular} \\[-5px]
    \scriptsize{(a) Kernel choices} & \scriptsize{(b) Input smoothing $s$}
    \vspace{-10px}
\end{tabular} 
\caption{\small
    \textbf{Model choices:}
    (a) Kernel choice on \textsl{50 Salads} \cite{stein2013combining}. 
    The combination of the Gaussian kernel and NTK, as given by the \eq{kern}, is the most accurate.
    (b) Input smoothing $\sigma$. 
    We consider three datasets with varying video lengths \textsl{50 Salads}\cite{stein2013combining} -- long, \textsl{Desktop Assembly} \cite{kumar2022unsupervised} -- medium, and \textsl{YTI} \cite{alayrac2016unsupervised} -- short. 
    The choice of the smoothing hyperparameter $s$ has effect, especially on long videos. 
}
\vspace{-15px}
\label{fig:abl}
\end{figure}

\medskip\noindent\textbf{Effect of the kernel choice.}
To test the effect of the kernel function, we consider seven kernel  options: 
\textsl{NNGP} -- the neural network Gaussian Process kernel \cite{lee2018deep} which is a characteristic kernel, 
\textsl{NTK} -- the neural tangent kernel \cite{arora2019exact} which is not characteristic,
\textsl{NTK sphere} -- the neural tangent kernel with inputs projected on the $\mathbb{S}^{d-1}$ sphere, where $d$ is the dimension of the inputs \cite{geifman2020similarity}, 
\textsl{SNTK} -- our implementation of the shift-invariant \textsl{NTK} proposed in \cite{jia2021efficient}, 
\textsl{Gauss} -- the simple Gaussian kernel with fixed length-scale, which is also characteristic, 
\textsl{Gauss${\times}$NTK sphere} -- using \eq{kern} to combine the \textsl{NTK sphere} and the Gaussian kernel, and 
\textsl{Gauss${\times}$NTK (Ours)} -- the combination of the Gaussian kernel and the \textsl{NTK} using \eq{kern}.
For the \textsl{Gauss} kernel we set the length-scale parameter to $\lambda{=}1e{+}1$, as that gave the best results, here. 
\fig{abl}(a) shows that while the characteristic kernels such as \textsl{Gauss}, \textsl{NNGP} and \textsl{NTK sphere} have a higher $F_1$ and $IoU$ scores than the rest, 
the combination of the Gaussian kernel with the NTK achieves the best segmentation accuracy. 

\medskip\noindent\textbf{Effect of input smoothing.}
Here, we test the effect of feature smoothing over time, on three datasets with varying video sizes: \textsl{50 Salads} (${\approx} 10$K frames) \cite{stein2013combining}, \textsl{Desktop Assembly} (${\approx} 700$ frames ) \cite{kumar2022unsupervised} and \textsl{YTI} \cite{alayrac2016unsupervised} (${\approx} 500$ frames).
We smooth the videos over time with a Gaussian filter over window sizes of approximately $s\frac{N}{M}$ and we vary the smoothing hyperparameter $s{\in}\{0.5, 1,1.5,2,2.5,3,3.5\}$.
In \fig{abl}(b) smoothing has a pronounced effect in terms of \textsl{MoF} scores, especially for long videos, as in \textsl{50 Salads}. 
This is because smoothing adds information about the neighboring actions.
In all experiments we set $s{=}2.5$ for datasets containing long videos (\ie \eg \textsl{50 Salads}, \textsl{MPII cooking}, \textsl{Breakfast}), and set $s{=}1.5$ for datasets containing short videos (\ie \textsl{YTI}, \textsl{Hollywood Extended} and \textsl{Desktop Assembly}).

\footnotetext[1]{
\scriptsize
Our implementation of \textsl{SNTK} proposed in \cite{jia2021efficient}.
}

\subsection{Unsupervised action segmentation per video}
\noindent\textbf{Average number of classes for segmentation.}
We compare our method with \textsl{all prior work} we could find, performing unsupervised action segmentation \textsl{per video} and reporting on these standard benchmarks.
Our method and the considered baselines segment each video independently, without relying on contextual information from large video datasets. 
We urge the reader to keep this setting in mind, while judging the completeness of our benchmarking. 
We do not add \textsl{SaM} \cite{xing2024unsupervised} because although their method is strikingly similar to \textsl{ABD} \cite{du2022fast}, they report results using the ground truth number of segments, while standardly the average number of segments per dataset\slash activity is used \cite{sarfraz2021temporally,xu2024temporally}.

\begin{table*}[t]
    \centering
    \resizebox{1\linewidth}{!}{
    \begin{tabular}{l ccc ccc ccc}
        \toprule
                & \multicolumn{3}{c}{Hollywood Extended} 
                & \multicolumn{3}{c}{Breakfast}  
                & \multicolumn{3}{c}{MPII Cooking 2}  \\
        \cmidrule(lr){2-4}
        \cmidrule(lr){5-7}
        \cmidrule(lr){8-10}
        & MoF ($\uparrow$) &  IoU ($\uparrow$) & $F_1$ ($\uparrow$)  
        & MoF ($\uparrow$) &  IoU ($\uparrow$) & $F_1$ ($\uparrow$)   
        & MoF ($\uparrow$) &  IoU ($\uparrow$) & $F_1$ ($\uparrow$)  \\ \midrule      
        TW-FINCH \cite{sarfraz2021temporally} (CVPR'21)  
            & 55.0\% & 35.0\% & 45.7\% 
            & 62.7\% & 42.3\% & 49.8\% 
            & 41.5\% & 22.4\% & 28.1\% 
            \\
        ABD \cite{du2022fast} (CVPR'22)     
            & 60.7\% & 36.0\% & 57.1\% 
            & 64.0\% &  ---   & 52.3\%  
            & ---    & ---    & --- 
            \\
        ASOT \cite{xu2024temporally} (CVPR'24)           
            & ---    & ---    & ---
            & 63.3\% & 35.9\% & 53.5\% 
            & ---    & ---    & ---
            \\
        CLOT \cite{bueno2025clot} (ICCV'25)
            & ---       & ---       & ---
            & 66.3\%    & 37.1\%    & 55.9\%
            & ---       & ---       & ---
        \\[2px]
        \cdashline{1-10}
        Ours (uniform init)
            & 54.1\% & 35.2\% &	46.4\% 
            & 57.9\% & 39.5\% &	47.5\% 
            & 35.3\% & 19.3\% & 25.1\% 
            \\
        Ours                
            & 55.1\% & 35.5\% &	46.5\%   
            & 61.5\% & 42.3\% &	49.8\%   
            & 39.7\% & 19.1\% &	25.4\% \\ 
        \toprule
                & \multicolumn{3}{c}{YTI} 
                & \multicolumn{3}{c}{Desktop Assembly}  
                & \multicolumn{3}{c}{50 Salads (Mid)} \\
        \cmidrule(lr){2-4}
        \cmidrule(lr){5-7}
        \cmidrule(lr){8-10}
        & MoF ($\uparrow$) &  IoU ($\uparrow$) & $F_1$ ($\uparrow$)   
        & MoF ($\uparrow$) &  IoU ($\uparrow$) & $F_1$ ($\uparrow$)  
        & MoF ($\uparrow$) &  IoU ($\uparrow$) & $F_1$ ($\uparrow$)  \\ \midrule      
        TW-FINCH \cite{sarfraz2021temporally} (CVPR'21)\footnotemark[1]  
            & 72.5\%    & 46.0\% & 58.2\%     
            & 73.3\%    & 57.8\% & 67.7\%     
            & 66.7\%    & 47.3\% & 57.1\% \\
        ABD \cite{du2022fast} (CVPR'22)\footnotemark[2]     
            & 67.2\%    & ---       & 49.2\%    
            & ---       & ---       & --- 
            & 71.8\%    & ---       & 59.2\%     \\
        ASOT \cite{xu2024temporally} (CVPR'24)           
            & 71.2\%    & 47.8\% & 63.3\%    
            & 73.4\%    & 47.6\% & 68.0\%    
            & 64.3\%    & 33.4\% & 51.1\%\\
        CLOT \cite{bueno2025clot} (ICCV'25)
            & 69.3\%    & 48.2\%    & 60.8\%    
            & 73.5\%    & 52.4\%    & 75.2\%
            & 69.4\%    & 45.0\%    & 63.8\%
        \\[2px]
        \cdashline{1-10}
        Ours (uniform init) 
            & 70.1\% & 45.1\% &	56.0\% 
            & 67.9\% & 48.4\% &	59.5\% 
            & 65.2\% & 46.0\% &	56.7\% 
        \\
        Ours 
            & 71.3\% & 46.0\% &	56.7\%   
            & 71.0\% & 52.4\% &	62.8\%
            & 70.1\% & 44.9\% & 55.3\% \\
        \bottomrule
    \end{tabular}}
\caption{\small
    \textbf{Benchmarking: average number of classes for segmentation.} 
    Results on six standard benchmarks compared to all baselines, performing unsupervised video segmentation per video.
    While \textsl{TW-FINCH} remains a strong baseline, our method shows competitive results, comparing positively to \textsl{ASOT} (CVPR'24) and \textsl{CLOT} (ICCV'25) on \textsl{50 Salads} and \textsl{ABD} (CVPR'22) on \textsl{YTI}.
}
\vspace{-20px}
\label{tab:exp3}
\end{table*}

\tab{exp3} shows the results across all datasets.
Overall, there is no winning method across all datasets.
Interestingly, \textsl{TW-FINCH} remains a strong baseline, especially for the datasets containing a high granularity level of annotations, like \textsl{YTI} and \textsl{Desktop Assembly}.
Our method obtains competitive results, outperforming \textsl{ASOT} \cite{xu2024temporally} on \textsl{50 Salads}. 
Our method cannot perform well on the \textsl{Desktop Assembly} because of the high granularity annotation level: \eg \textsl{"tighten\_screw1"}, \textsl{"tighten\_screw2"}, \textsl{"tighten\_screw3"}, \textsl{"tighten\_screw4"}.
On the \textsl{Hollywood Extended} and \textsl{Breakfast} we only learn 3 and 6 synthetic frames in the video approximation, respectively. 
Such a low number of synthetic frames may not be sufficient to learn a good video approximation.
On the datasets with short videos, such as \textsl{YTI}, \textsl{Hollywood Extended} our method gains little in learning the video approximation on top of the uniform initialization. 
This may indicate that the models are overfitting here.
On the other hand, we observe a large gain in segmentation accuracy (MoF) for long videos such as \textsl{50 Salads} and \textsl{MPII Cooking 2}, when compared to the uniform initialization.
On average, our method learns a video approximation that align well with the original video in the deep kernel space.

\footnotetext[1]{
\scriptsize
Where missing, we report \textsl{TW-FINCH}\cite{sarfraz2021temporally} results as run by us. 
}
\footnotetext[2]{
\scriptsize
\textsl{ABD}\cite{du2022fast} do not provide code and we failed to reproduce their results.
}

\smallskip\noindent\textbf{Random number of classes for segmentation.}
\begin{table*}[t]
    \centering
    \resizebox{.9\linewidth}{!}{
        \begin{tabular}{l ccc ccc ccc}
        \toprule
                & \multicolumn{3}{c}{Hollywood Extended} 
                & \multicolumn{3}{c}{Breakfast}  
                & \multicolumn{3}{c}{MPII Cooking 2} \\
        \cmidrule(lr){2-4}
        \cmidrule(lr){5-7}
        \cmidrule(lr){8-10}
    & MoF ($\uparrow$) &  IoU ($\uparrow$) & $F_1$ ($\uparrow$)   
    & MoF ($\uparrow$) &  IoU ($\uparrow$) & $F_1$ ($\uparrow$)  
    & MoF ($\uparrow$) &  IoU ($\uparrow$) & $F_1$ ($\uparrow$)  \\ \midrule      
         TW-FINCH \cite{sarfraz2021temporally}(CVPR'21)  
            & 56.4\% & 33.0\% & 36.7\%  
            & 58.4\% & 41.3\% & 38.0\% 
            & 40.5\% & 24.4\% & 21.9\% \\
        Ours
            & 53.3\% & 34.7\% & 39.3\% 	 
            & 57.9\% & 41.4\% & 38.6\% 
            & 39.3\% & 22.3\% & 20.8\%\\
   \end{tabular}}
    \resizebox{.9\linewidth}{!}{
    \begin{tabular}{l ccc ccc ccc}
        \toprule
                & \multicolumn{3}{c}{YTI} 
                & \multicolumn{3}{c}{Desktop Assembly}  
                & \multicolumn{3}{c}{50 Salads (Mid)} \\
        \cmidrule(lr){2-4}
        \cmidrule(lr){5-7}
        \cmidrule(lr){8-10}
    & MoF ($\uparrow$) &  IoU ($\uparrow$) & $F_1$ ($\uparrow$) 
    & MoF ($\uparrow$) &  IoU ($\uparrow$) & $F_1$ ($\uparrow$)  
    & MoF ($\uparrow$) &  IoU ($\uparrow$) & $F_1$ ($\uparrow$)  \\ \midrule      
     TW-FINCH \cite{sarfraz2021temporally}(CVPR'21)  
            & 67.3\% & 44.5\% & 47.3\%  
            & 60.8\% & 46.9\% & 42.9\%  
            & 57.2\% & 41.3\% & 38.2\% \\
        Ours 
            & 66.5\% & 43.6\% & 47.5\%   
            & 61.2\% & 41.8\% & 43.1\% 	 
            & 60.3\% & 38.7\% & 41.8\%	\\ 
     \bottomrule
    \end{tabular}}\\[5px]
    \begin{tabular}{c@{\hskip 0.2in}c}
    \includegraphics[width=.45\linewidth, decodearray={0.5 1 0.5 1 0.5 1}]{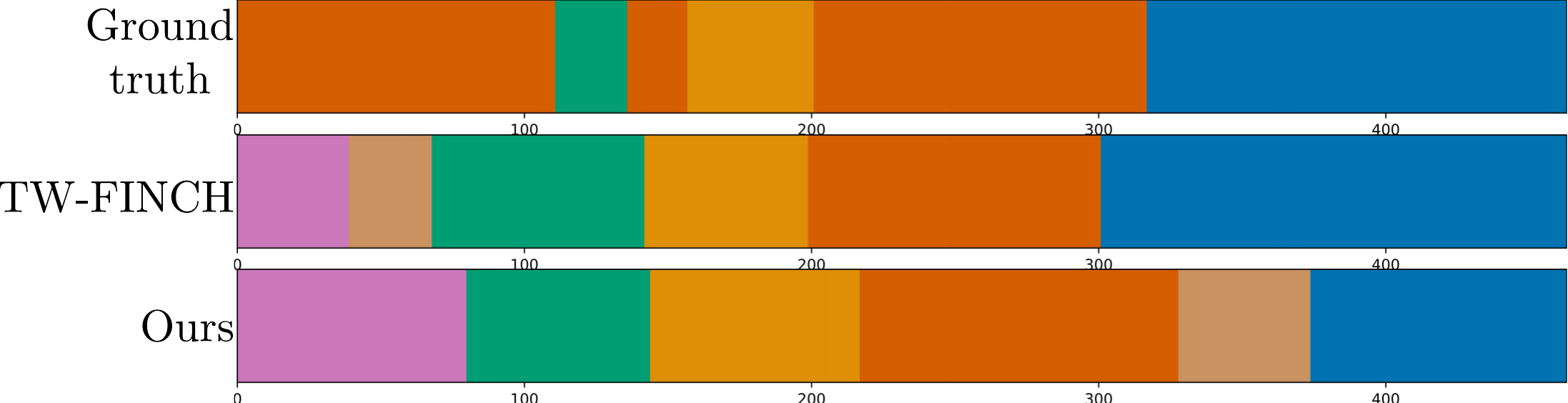} &
    \includegraphics[width=.45\linewidth, decodearray={0.5 1 0.5 1 0.5 1}]{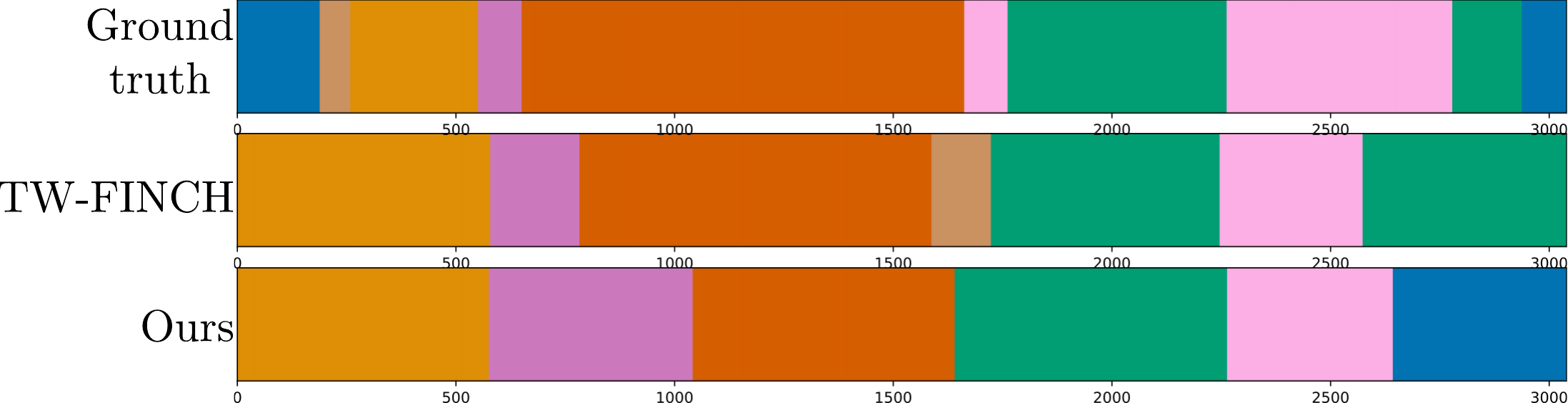} \\
    \scriptsize{$5^{th}$ video example from \textsl{Hollywood Extended}} &
    \scriptsize{$5^{th}$ video example from \textsl{Breakfast}} \\
    \includegraphics[width=.45\linewidth, decodearray={0.5 1 0.5 1 0.5 1}]{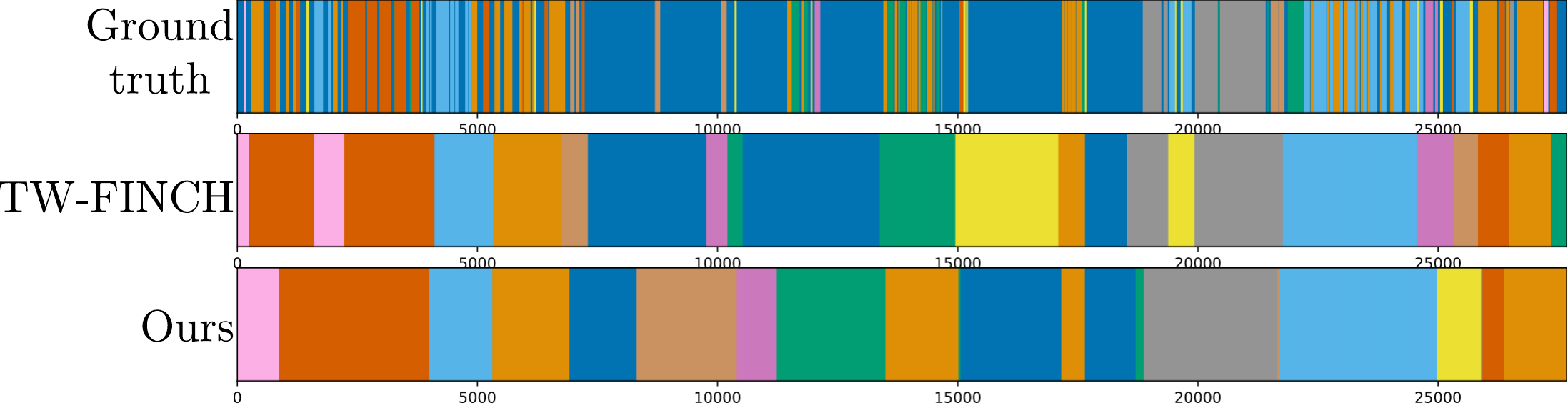} &
    \includegraphics[width=.45\linewidth, decodearray={0.5 1 0.5 1 0.5 1}]{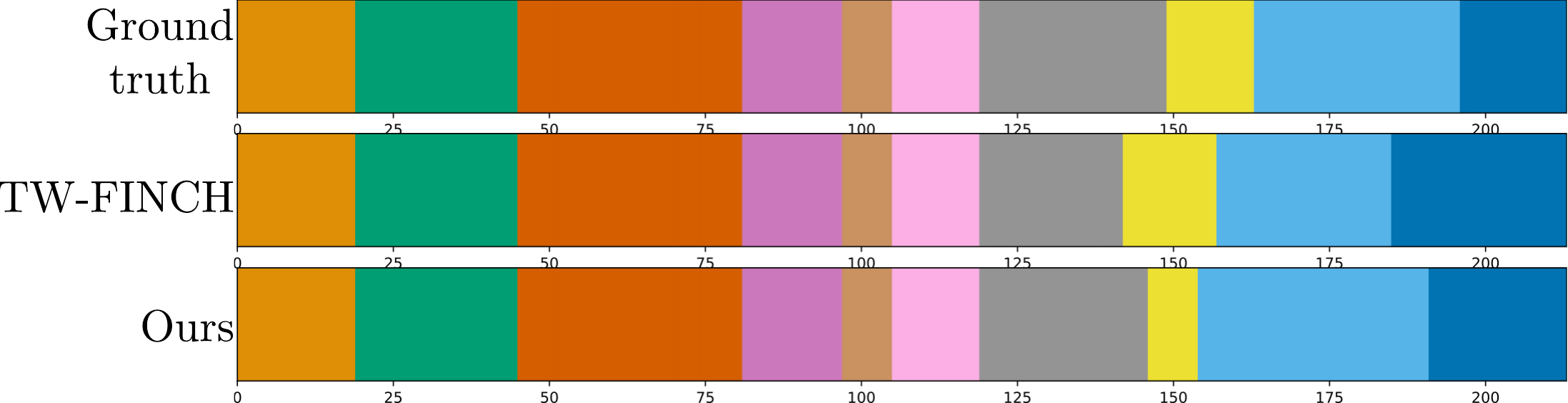} \\
    \scriptsize{$5^{th}$ video example from \textsl{MPII Cooking 2}} &
    \scriptsize{$5^{th}$ video example from \textsl{YTI}} \\
    \includegraphics[width=.45\linewidth, decodearray={0.5 1 0.5 1 0.5 1}]{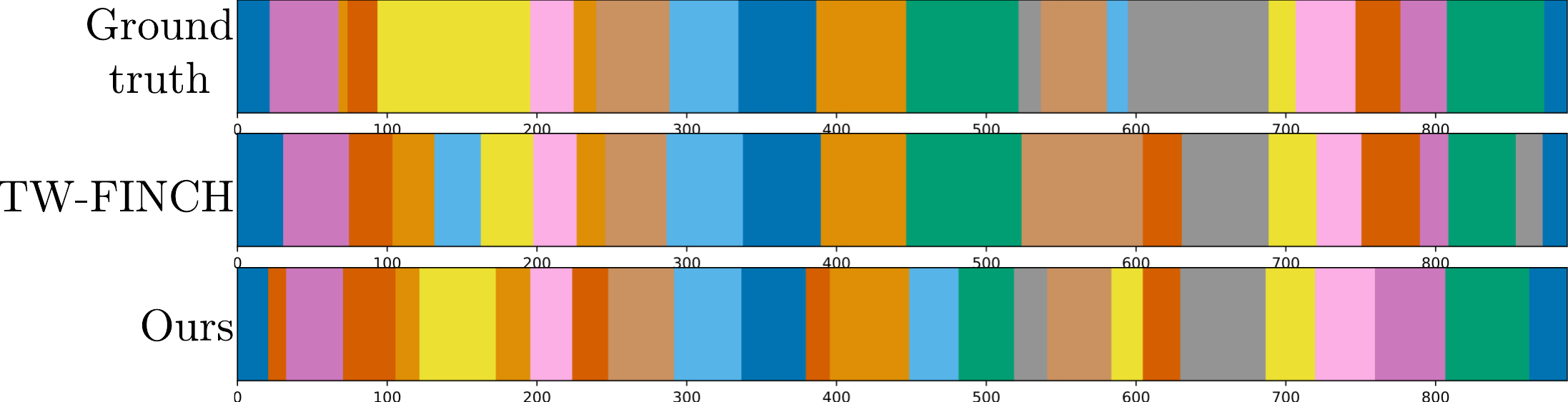} &
    \includegraphics[width=.45\linewidth, decodearray={0.5 1 0.5 1 0.5 1}]{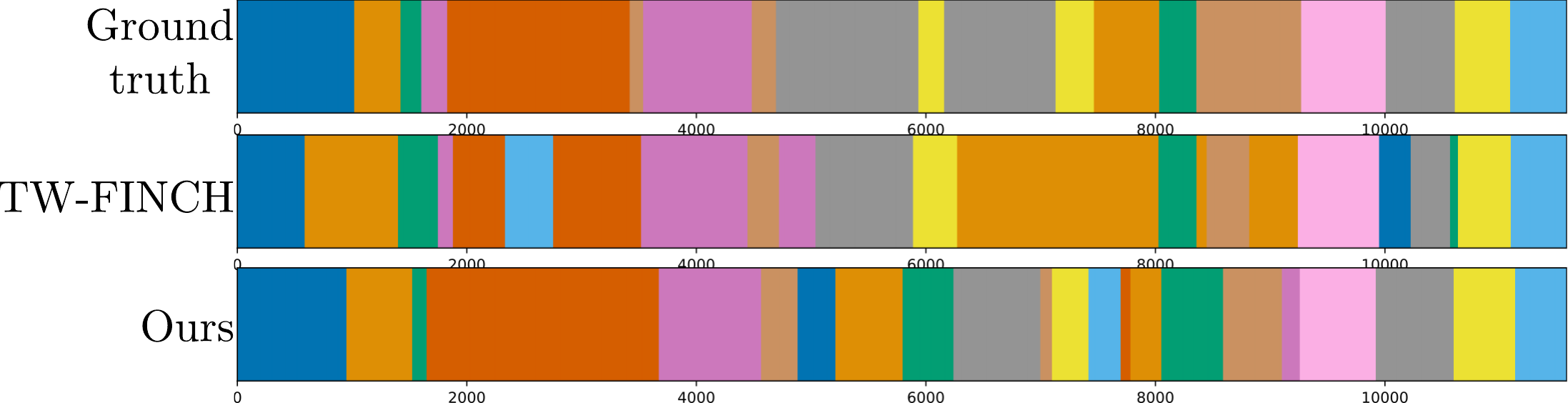} \\
    \scriptsize{$5^{th}$ video example from \textsl{Desktop Assembly}} &
    \scriptsize{$5^{th}$ video examples from \textsl{50 Salads (Mid)}} \\
    \end{tabular}
    \caption{\small
    \textbf{Benchmarking: random number of classes for segmentation.}
    Our proposed method repeatedly outperforms the \textsl{TW-FINCH} baseline in terms of $F_1$ scores, when the ideal number of segments is not known, which is a realistic setting.
    We also show example predictions for the $5^{th}$ video of each dataset.
    }
    \label{tab:exp3b}
    \vspace{-20px}
\end{table*}
Here, we test the hypothesis that hierarchical clustering creates arbitrary segment boundaries that do not respect the underlying video data.
This is a realistic setting, as often the true underlying number of segments present in the video is not known. 
For this, we rerun our code and the code of FW-FINCH \cite{sarfraz2021temporally} on all datasets using $M {=} \overline{M} {\pm} \mathcal{U}({-}\overline{M},\overline{M})$, where $\overline{M}$ is the average number of actions annotated in each dataset.
We use the same random seed for both methods.
Similar to the synthetic data, to better deal with the noisy number of segments, we train for 200 iterations with a weight decay of $1e{-}4$.
We cannot compare with ABD \cite{du2022fast} as the code is not published and our efforts to replicate the author's results did not succeed.
We do not consider ASOT \cite{xu2024temporally} and CLOT \cite{bueno2025clot} as they do not do hierarchical clustering.   

\tab{exp3b} shows the results when the ideal number of segments is not known.
While \textsl{TW-FINCH} shows competitive results in \tab{exp3} when the number of segments is on average the true one, it is less competitive when the required number of segments largely deviates from the true one. 
Our method consistently outperforms the \textsl{TW-FINCH} baseline in $F_1$ scores, except for the \textsl{MPII Cooking 2} dataset, where our smooth predictions cannot detect the dense background.
This improvement is due to our method not forcing a fixed number of segments. 
By assigning each frame to its closest approximation in kernel space, it can happen that some synthetic frames do have any video frames assigned to them. 
Therefore, our method more closely follows the underlying data distribution when segmenting the videos.
We also illustrate under \tab{exp3b} a randomly chosen example (the predictions for the 5$^{th}$ video) from each dataset.

%% file: paper/discussion.tex
\section{Method limitations and conclusions}
\begin{figure}[t]
    \centering
    \begin{tabular}{cc}
    \includegraphics[width=.5\linewidth, decodearray={0.5 1 0.5 1 0.5 1}]{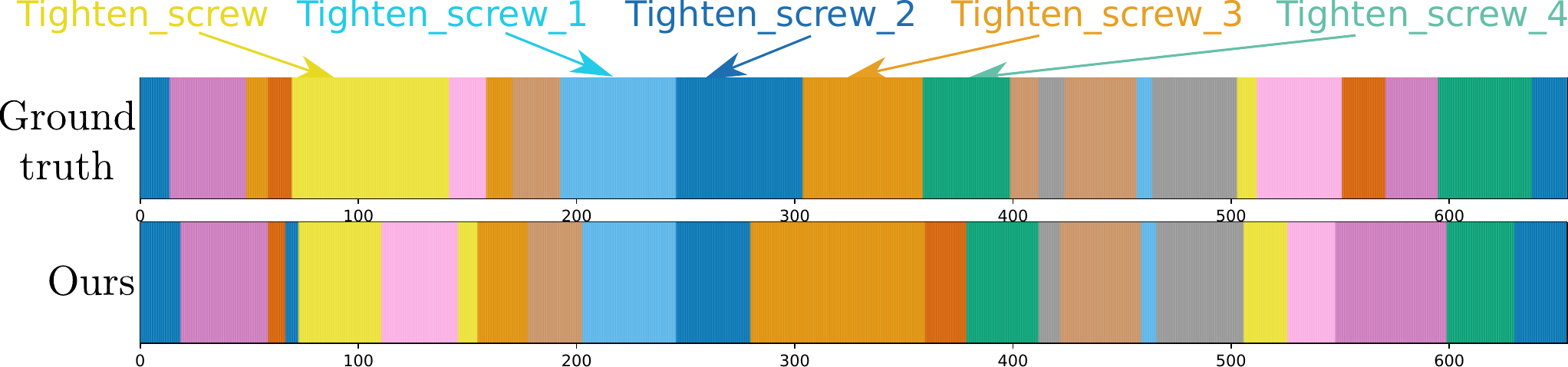} &
    \includegraphics[width=.5\linewidth, decodearray={0.5 1 0.5 1 0.5 1}]{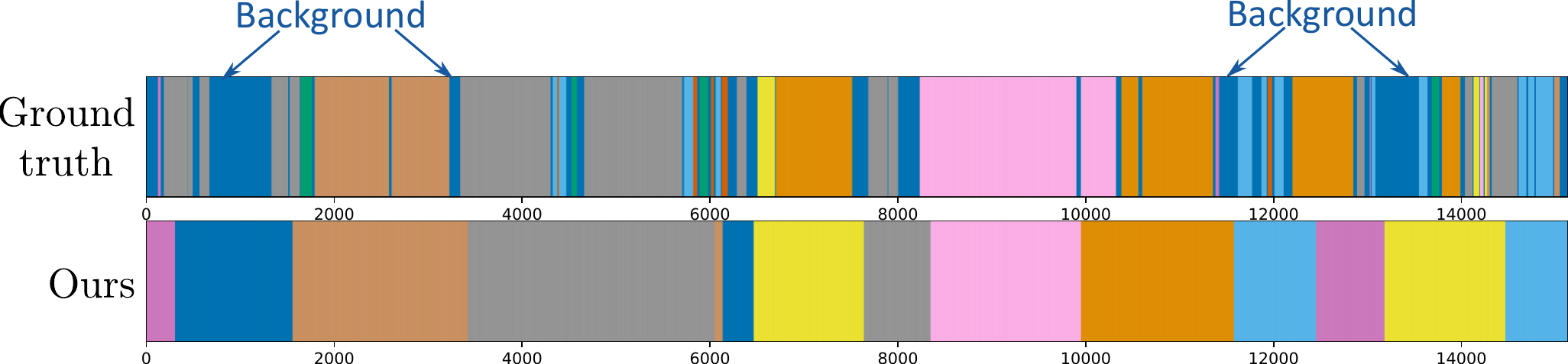} \\
    \small{\scriptsize (a) Failures on \textsl{Desktop Assembly} } &
    \small{\scriptsize (b) Failures on \textsl{MPII Cooking 2} }\\
    \small{\scriptsize (video \textsl{2020-06-03\_18-01-32}) } &
    \small{\scriptsize (video \textsl{s08-d04-cam-002}) }\\
    \end{tabular}
    \caption{\textbf{Failure cases.} 
    Our method fails to correctly segment very similar actions, as in the case of \textsl{Desktop Assembly}.
    Another failure is when distinct actions are grouped together, such as the background class on the challenging \textsl{MPII Cooking 2} dataset.
    }
    \vspace{-15px}
    \label{fig:fail}
\end{figure}
While being effective at learning video approximations, especially for long videos, our method\rq s descriptive power is bounded. 
Concretely, our method learns a single synthetic frame per action, which may be over-simplistic for long and complex actions.
Another limitation of our proposed method is that it fails to correctly segment the video if the actions are too similar, as in the case of \textsl{Desktop Assembly} (\eg \textsl{"tighten\_screw\_1"}, \textsl{"tighten\_screw\_2"}, \etc) in \fig{fail}(a).
Another failure case is when actions that are dissimilar are annotated together under a single name, \eg the background action in \textsl{MPII Cooking 2}, as in \fig{fail}(b).
Additionally, our method is sensitive to the smoothing hyperparameter, $s$. 
Learning the smoothing hyperparameter together with the video approximation is possible, however this cannot be directly used in combination with batched data.
And batched data is, in practice, better than using the complete video.

To conclude, we propose a simple method for learning video approximations and aligning these with the original video in kernel space, for unsupervised action segmentation.
We show that the proposed method works in practice, and that the learned video approximation aligns well with the original video.
Our method has the added benefit that it respects the underlying structure of the data when the true number of actions is not known.

\medskip\noindent\textbf{Acknowledgements.} This work is supported by the IWISH project (ITEA 20044) and funded by RVO (Netherlands Enterprise Agency).